\documentclass[runningheads]{llncs}
\usepackage[T1]{fontenc}
\usepackage[colorlinks,bookmarksopen,bookmarksnumbered,citecolor=blue, linkcolor=blue, urlcolor=blue]{hyperref}
\usepackage{placeins}
\usepackage{enumitem}
\usepackage{amsmath}
\usepackage{amsfonts}
\usepackage{multirow}
\usepackage{booktabs}
\usepackage{placeins}
\usepackage{textcomp}
\usepackage{makecell}
\usepackage{marvosym}
\newcommand{\corresponding}{\textsuperscript{\Letter}}
\usepackage{graphicx,verbatim}
\begin{document}
\title{KANResDiff: Learning Local Residual Diffusion via Kolmogorov-Arnold Network for Ambiguous Medical Image Segmentation}
\titlerunning{KANResDiff for Ambiguous Medical Image Segmentation}
%
%
\author{Fanding Li\inst{1} \and Chenglin Wang\inst{1} \and Xiangyu Li\inst{1}\corresponding \and Xingyu Qiu\inst{1} \and Xinghua Ma\inst{1} \and Xiangming Yin\inst{1} \and Haiyang Li\inst{1} \and Suyu Dong\inst{2} \and Wei Wang\inst{3} \and Kuanquan Wang\inst{1} \and Gongning Luo\inst{1} \and Shuo Li\inst{4}}


\authorrunning{F. Li et al.}
%
\institute{Faculty of Computing, Harbin Institute of Technology, Harbin, China \and College of Computer and Control Engineering, Northeast Forestry University, Harbin, China \and Faculty of Computing, Harbin Institute of Technology, Shenzhen, China \and Department of Computer and Data Science and Department of Biomedical Engineering, Case Western Reserve University, Cleveland, Ohio 44106, United States\\
\email{lixiangyu@hit.edu.cn}}
\maketitle              
\begin{abstract}
Ambiguous medical image segmentation aims to provide a series of diverse but plausible segmentation hypotheses. However, existing methods introduce stochasticity in a fixed and pre-defined manner, failing to form a progressive semantic modeling process. To address these challenges, we propose KANResDiff to learn local residual diffusion with Kolmogorov-Arnold Network, thereby assigning distinct roles across stages for ambiguity modeling. Specifically, we propose Independent Time Encoding that offers spline-based time embeddings instead of linear ones from MLPs, which enhances the independence across inference stages and assigns progressive semantic roles to different stages. We propose Residual Schr\"odinger Bridge that injects deterministic residual prior with learnable weights by constructing local Schr\"odinger Bridge instead of following manually settings, achieving a flexible deterministic–stochastic interaction and stage-aware ambiguity modeling thanks to local optimal diffusion path. Extensive experimental results on two public datasets demonstrate that KANResDiff achieves SOTA performance on GED and HM-IoU, with maximum improvements of $16.8\%$ and $7.7\%$, respectively, while maintaining competitive performance on the MDM metric. Source code is available at \url{https://github.com/PerceptionComputingLab/KANResDiff}.
	
\keywords{Ambiguous Medical Image Segmentation  \and  Diffusion Model \and Kolmogorov-Arnold Network \and Schr\"odinger Bridge.}
	
\end{abstract}
\section{Introduction}
Ambiguous medical image segmentation (AMIS), aiming to generate a series of diverse yet anatomically consistent predictions that reflect the underlying annotation distribution \cite{li2025ambiguity,wang2025contour}, relies on the explicit introduction of stochasticity during inference to capture the inherent variability in clinical annotations. By exploring multiple anatomically consistent solutions, stochastic inference provides a more reliable and comprehensive basis for clinical decision-making, improving robustness against ambiguous boundaries and inter-observer variability \cite{kohl2018probabilistic,rahman2023ambiguous}.

However, existing AMIS methods introduce stochasticity in a fixed and inflexible manner during inference, leading to improper stochastic intensity, where randomness is either over-injected or under-explored. Existing methods can be broadly categorized into three groups: cVAE-based models \cite{kohl2018probabilistic,baumgartner2019phiseg}, logit distribution-based models \cite{monteiro2020stochastic,zhang2022pixelseg}, and diffusion models \cite{li2025ambiguity,zbinden2023stochastic}. In the former two categories, ambiguity is introduced only at the final prediction stage, whereas in the latter, stochasticity is incorporated throughout the entire inference trajectory. Consequently, existing methods are constrained by fixed stochastic formulations, failing to establish an AMIS-specific progressive semantic modeling process.

\begin{figure} [t]	
	\centering 	 	 
	\includegraphics[width = \textwidth]{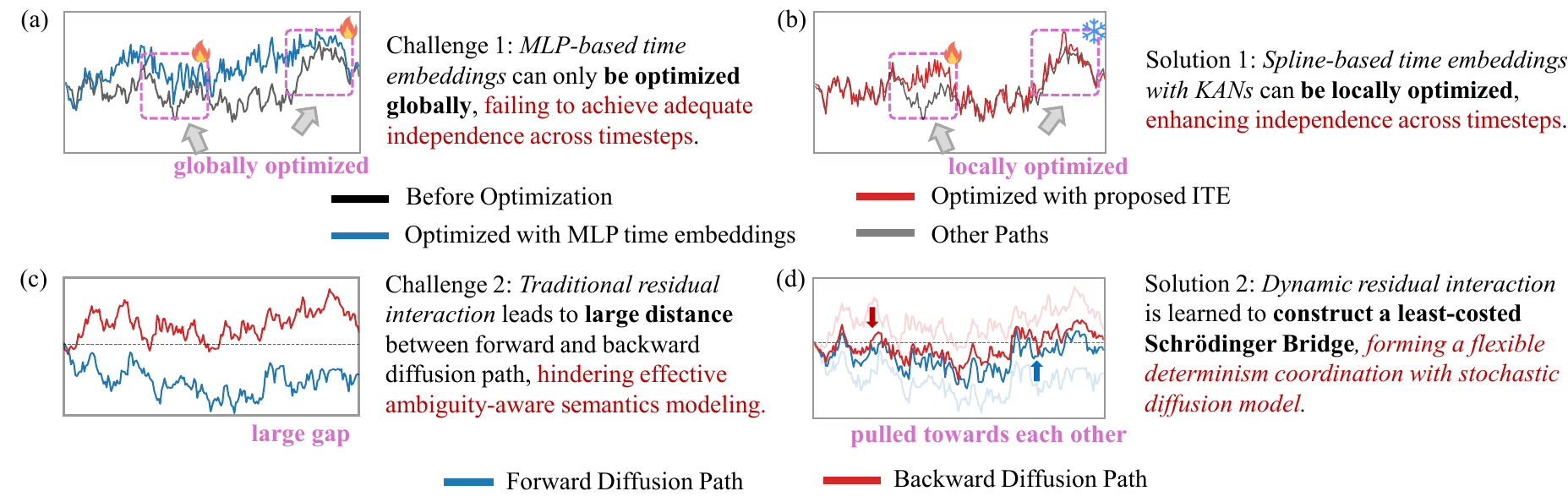} 	
	\caption{(a) Traditional MLP-based time embeddings can only be optimized globally. (b) Our proposed KANResDiff enhances indepedence across timesteps via spline-based time embeddings, which can be locally optimized. (c) Traditional residual diffusion leads to large gap between forward and backward diffusion path, hindering effective semantics modeling. (d) Our proposed KANResDiff constructs local Schr\"odinger Bridge with dynamic residual components to form a least-costed diffusion path, where forward and backward diffusion paths are close to each other.} 
	\label{fig:intro}
\end{figure}

Residual-based diffusion \cite{han2022card,shang2024resdiff} has demonstrated potential in alleviating dynamic stochasticity injection thanks to the mechanism that explicitly involves deterministic predictors into diffusion formulation. Residual-based diffusion enables a unified inference mechanism in which the relative roles of determinism and stochasticity can be explicitly controlled, which is widely used in multiple areas such as classification, regression and image super-resolution.

However, directly applying Residual-based diffusion for dynamic and progressive stochasticity injection for AMIS remains challenging due to: \textbf{(1) Insufficient independence among inference states hinders the ability to represent heterogeneous ambiguity and undermine effective ambiguity-aware inference in AMIS. (Fig. \ref{fig:intro}(a))} Residual-based diffusion typically represents inference stages using time embeddings parameterized by MLPs, which induces strong parameter coupling and limited effective independence across inference steps. While such a design is generally sufficient for holistic generative tasks such as classification \cite{han2022card}, regression \cite{han2022card}, and image super-resolution \cite{shang2024resdiff}, it becomes insufficient for AMIS, where ambiguity must be progressively constructed along inference, requiring greater independence across intermediate stages. 
\textbf{(2) Stage-agnostic residual interaction constrains the flexibility of deterministic–stochastic coordination, limiting effective ambiguity-aware inference in AMIS. (Fig. \ref{fig:intro}(c))} Residual-based diffusion typically introduces deterministic predictors through a fixed residual formulation, where the relative contribution between deterministic predictors and stochasticity in diffusion is predefined across inference stages. However, inference stages play different roles in ambiguity modeling in AMIS: early stages emphasize structural consistency and later stages express stochasticity to capture diverse plausible segmentations, requiring dynamic and stage-specific residual injection.

To address the aforementioned challenges, we propose KANResDiff to learn local residual diffusion via Kolmogorov-Arnold Networks (KANs) for a flexible deterministic–stochastic coordination with both better independence across inference stages and dynamic, stage-specific residual injection. Specifically, 
(1) We propose \textbf{Independent Time Encoding}, a time reparameterization with localized function representations to reduce cross-stage interference during diffusion inference, enhancing \textbf{local independence across inference stages}. Specifically, ITE models the temporal variable with locally supported basis functions, such that optimizations at one inference stage minimally affect others, thereby enhancing step-wise independence and assigning progressive ambiguity modeling across timesteps. (Fig. \ref{fig:intro}(b))
(2) We propose \textbf{Residual Schr\"odinger Bridge (RSB)}, an AMIS-specific residual interaction strategy by constructing Schr\"odinger Bridge at each inference stages, assigning \textbf{local residual weights}. Specifically, RSB assigns step-specific residual weights to diffusion formulations by constructing Schr\"odinger Bridge across timesteps, ensuring an optimal stochastic inference by intruducing dynamic residual deterministic components, thereby achieving a flexible determinism injection for diffusion in AMIS. (Fig. \ref{fig:intro}(d))

In summary, our main contributions are as follows:
\begin{itemize}
	\item We propose KANResDiff to learn local residual diffusion, the first paradigm that enables dynamic and progressive stochastic modeling for ambiguous medical image segmentation, thereby establishing a principled semantic modeling trajectory for ambiguity representation.
	\item The proposed ITE introduces a novel time-embedding reparameterization strategy based on locally supported basis functions, which significantly enhances the independence across different timesteps and facilitates a progressive ambiguity modeling process.
	\item The proposed RSB introduces a novel residual guidance mechanism by constructing a local Schrödinger Bridge, which dynamically regulates the contribution of residual deterministic components, thereby enabling flexible and adaptive determinism injection into the diffusion process for AMIS.
	\item Experimental results on the LIDC dataset and ISIC subset demonstrate that KANResDiff outperforms existing methods across evaluation metrics.
\end{itemize}

\section{Methods}

\begin{figure} [t]	
	\centering 	 	 
	\includegraphics[width = \textwidth]{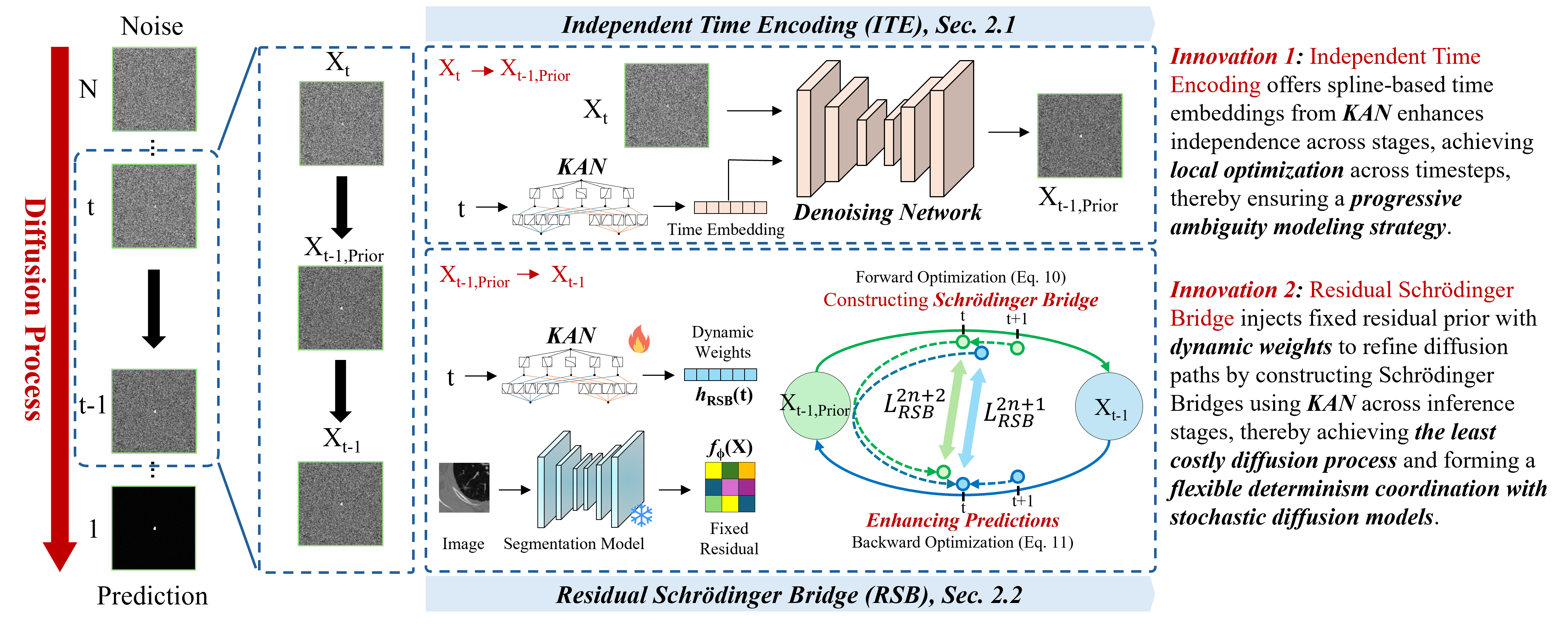} 	
	\caption{The proposed KANResDiff achieves superior performance via a flexible deterministic–stochastic interaction in diffusion models by enhancing independence across inference steps (ITE, Sec. 2.1) and assigning dynamic residual weights to construct local Schr\"odinger Bridges (RSB, Sec. 2.2).} 
	\label{fig:main}
\end{figure}

The proposed KANResDiff (Fig. \ref{fig:main}) achieves a flexible deterministic–stochastic interaction in diffusion models via Kolmogorov-Arnold Networks from two aspects: 
Firstly, a time embedding strategy with locally supported basis functions enhances independence across timesteps, assigning a progressive ambiguity modeling role to each timestep (\textit{ITE, Sec. 2.1}).
Secondly, local Schr\"odinger Bridges are constructed via introducing deterministic prior residual with dynamic stage-aware weights to diffusion formulations, achieving a flexible determinism injection to stochasticity of diffusion models, thereby forming a flexible deterministic–stochastic coordination in AMIS (\textit{RSB, Sec. 2.2}).

\subsection{Independent Time Encoding Constructs a Progressive Ambiguity Modeling Procedure.}

The proposed Independent Time Encoding (ITE) reparameterizes time embeddings with locally supported basis functions, enhancing local optimization independence across timesteps, thereby ensuring a progressive ambiguity modeling strategy along inference stages. 

Diffusion models approximate T consecutive denoising networks with a single network taking time embeddings from MLPs as input due to excessive memory.
Gradients from MLP-based time embeddings at timestep $t_1$ is:
\begin{equation}
	\frac{\partial L(t_1)}{\partial W}
	=
	\frac{\partial L}{\partial g(t_1)}
	\cdot
	\frac{\partial g(t_1)}{\partial W}
	\neq 0 .
\end{equation}
since parameter set $W$ is shared across all timesteps with MLP-based time embeddings, the embedding of any timestep $t_2 \neq t_1$ is optimized as eq. \ref{eq2}. Consequently, all time embeddings are modified with optimization at timestep $t_1$, causing inadequate independence across timesteps for AMIS:
\begin{equation}
	\label{eq2}
	\frac{\partial L(t_2)}{\partial g(t_1)}
	=
	\frac{\partial L(t_2)}{\partial W}
	\cdot
	\frac{\partial W}{\partial g(t_1)} \neq 0.
\end{equation}

\begin{figure} [t]	
	\centering 	 	 
	\includegraphics[width = \textwidth]{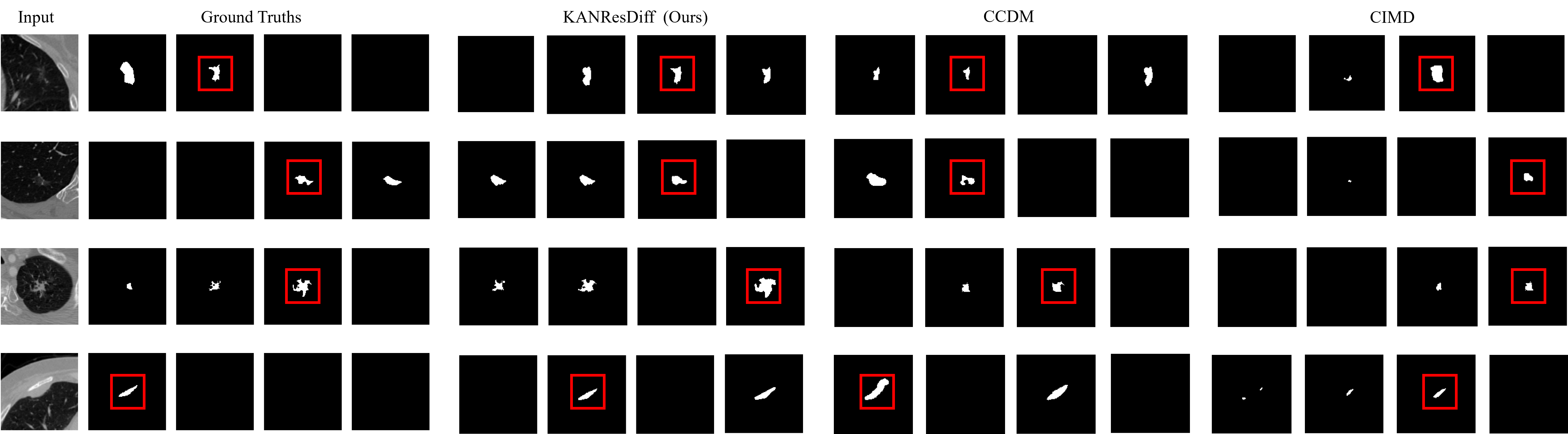} 	
	\caption{Comparative qualitative results on LIDC dataset among ground truths, two advanced methods and the proposed KANResDiff demonstrate both better alignment with ground truths and higher per-sample accuracy.} 
	\label{fig:pred}
\end{figure}

In our ITE, time embeddings are formulated with B-splines as eq. \ref{1} for independent optimization during training, enabling a progressive semantic accumulation for plausible ambiguity modeling:
\begin{equation}
	\label{1}
	g_{spline}(t) = \sum_{i} c_i \times B_{i,p}(t); B_{i,p}(t) = 0, \text{when} \  t \notin [t_i, t_{i+p+1})
\end{equation}
where $B_{i,p}(\cdot)$ denotes the i$^{\text{th}}$ B-spline basis function with degree p and $c_i$ are learnable coefficients. Therefore, for the optimization at $t_3 \in [t_i, t_{i+p+1})$, the gradient of loss function $L(t_3)$ is calculated as follows:
\begin{equation}
	\frac{\partial L(t_3)}{\partial c_i}
	=
	\frac{\partial L}{\partial g(t_3)}
	\cdot
	B_{i,p}(t_3)
\end{equation}
Consequently, if $t_4$ does not fall within the support interval of the B-spline basis functions for $t_3$, i.e., $t_4 \notin [t_i, t_{i+p+1})$, then
\begin{equation}
	\frac{\partial L(t_4)}{\partial c_i}
	=
	\frac{\partial L}{\partial g(t_3)}
	\cdot
	B_{i,p}(t_4)
	=
	0
\end{equation}
achieving local independent optimization with spline-based parameterizations, thereby ensuring a progressive ambiguity modeling strategy.

For implementation, we utilize cubic B-spline (i.e. $p=3$) that provides a balanced trade-off between smooth temporal evolution and local parameter updates. Kolmogorov-Arnold Network provides a principled instantiation of spline-based time encoding via its resistance to forgetting and continuous output, making it a suitable and necessary choice for implementation of ITE.

\subsection{Residual Schr\"odinger Bridge Forms a Flexible Determinism Coordination with Stochastic Diffusion Models.}

The proposed Residual Schr\"odinger Bridge (RSB) assigns dynamic weights instead of manually predefined weights to residual components by constructing local Schr\"odinger Bridges across inference stages, thereby achieving the least costly diffusion process and forming a flexible determinism coordination with stochastic diffusion models. Deterministic residual prior is significant in ambiguous medical image as a constraint of overall plausibility. Traditional residual diffusion models introduce residual components into inference via fixed formulation as eq. \ref{8} without maintaining the true anatomy of diffusion paths in latent space, failing to preserve the semantic plausibility modeling \cite{li2025ambiguity,qiu2025finding}.

\begin{equation}
	\label{8}
	x_{t} = \sqrt{(\bar{\alpha_t})} \times x_{0} + (1 - \sqrt{(\bar{\alpha_t})}) \times f_{\phi}(X) + \sqrt{1 - \bar{\alpha_t}} \times \epsilon
\end{equation}

To address the limitations, we optimize the diffusion path for AMIS by constructing Schr\"odinger Bridges across inference stages and regarding residual components as additional refinement to original stochastic diffusion path. Specifically, ambiguity-aware inference procedure of the proposed RSB is:
\begin{equation}
	\label{9}
	x_{t} = \sum_{i=1}^n \frac{\sqrt{\bar{\alpha_t}}}{n} \times x_{0i} + h_{\text{RSB}}(t) \times f_{\phi}(X) + \sqrt{1 - \bar{\alpha_t}} \times \epsilon
\end{equation}
where $h_{\text{RSB}}(t)$ is the dynamic weight at timestep t, serving both the refinement of original diffusion path to construct a Schr\"odinger Bridge and a flexible determinism injection maintaining overall plausibility. $n$ denotes the number of plausible predictions, making the proposed RSB an AMIS-effective approach.

\begin{table*}[t]      	
	\caption{Quantitative results on LIDC dataset show the superior performance of KANResDiff. \textbf{Bold} represents the best per column.}      	
	\label{tab:1}      	
	\centering      		
	\begin{tabular}{l|ccc|c|cc}          			
		\toprule          			
		Methods & $\text{GED}_{16} \downarrow$ & $\text{GED}_{32} \downarrow$ & $\text{GED}_{100} \downarrow$ &  $\text{HM-IoU}_{32} \uparrow$ & $\text{MDM}_{32}$ $\uparrow$ \\
		
		\midrule          			
		Prob. Unet\cite{kohl2018probabilistic} & $0.310_{\pm0.010}$ & $0.303_{\pm0.010}$ & $0.252_{\pm0.004}$ & $0.548_{\pm 0.000}$ & $0.681_{\pm0.020}$  \\  
		
		MoSE\cite{20} & $0.218_{\pm0.003}$ & $0.195_{\pm0.002}$ & $0.189_{\pm0.002}$ & $0.624_{\pm0.004}$ & $0.767_{\pm0.004}$ \\
		
		P$^{\text{2}}$SAM\cite{25} & $0.208_{\pm0.000}$ & $0.206_{\pm0.000}$ & $0.206_{\pm0.000}$ & $0.627_{\pm0.000}$ & $0.939_{\pm0.000}$ \\
		
		CIMD\cite{rahman2023ambiguous} & $0.234_{\pm0.005}$ & $0.218_{\pm0.005}$ & $0.210_{\pm0.005}$ & $0.592_{\pm0.002}$ & $0.915_{\pm0.004}$\\  
		
		AB\cite{19} & $0.213_{\pm0.001}$ & $0.196_{\pm0.002}$ & $0.193_{\pm0.002}$ & $0.619_{\pm0.001}$ & $0.792_{\pm0.002}$ \\ 
		
		CCDM\cite{zbinden2023stochastic} & $0.212_{\pm0.001}$ & $0.194_{\pm0.001}$ & $0.183_{\pm0.002}$ & $0.631_{\pm0.002}$ & $0.790_{\pm0.003}$ \\
		
		SSB\cite{baru2025ambiguous} & - & $0.245_{\pm0.003}$ & $0.208_{\pm0.001}$ & - & $\mathbf{0.942_{\pm0.001}}$ \\
		
		ContourMS\cite{wang2025contour} & $0.231_{\pm0.003}$ & $0.212_{\pm0.002}$ & $0.203_{\pm0.003}$ & $0.651_{\pm0.005}$ & - \\
		
		KANResDiff(Ours) & $\mathbf{0.187_{\pm0.003}}$ & $\mathbf{0.181_{\pm0.002}}$ & $\mathbf{0.172_{\pm0.001}}$ & $\mathbf{0.701_{\pm0.001}}$ & $0.930_{\pm0.002}$ \\  
		\bottomrule
		
	\end{tabular} 
\end{table*} 

To form a Schr\"odinger Bridge, the inference procedure in eq. \ref{9} should follow SB-formed forward and backward PDEs as eq. \ref{10} and \ref{11}, where the residual term $h_{\text{RSB}}(t) \times f_{\phi}(X)$ serves as the control drift for minimum cost optimization:
\begin{equation}
	\label{10}
	\mathrm{d}X_t
	=
	\Big(
	\sqrt{(\bar{\alpha_t})}X_t + h_{\mathrm{RSB}}(t)f_{\phi}(X)
	+ \beta(t)\nabla_x\log \Psi(X_t,t)
	\Big)\mathrm{d}t
	+
	\sqrt{\beta(t)}\,\mathrm{d}W_t
\end{equation}
\begin{equation}
	\label{11}
	\mathrm{d}X_t
	=
	\Big(
	\sqrt{(\bar{\alpha_t})}X_t + h_{\mathrm{RSB}}(t)f_{\phi}(X)
	- \beta(t)\nabla_x\log \Psi(X_t,t)
	\Big)\mathrm{d}t
	+
	\sqrt{\beta(t)}\,\mathrm{d}W_t.
\end{equation}

For implementation, we perform refinement based on a pretrained DDPM with fixed noise schedule. The forward and backward procedures of SB are iteratively optimized to be as similar as possible for each timestep t. Specifically, the optimization and training objectives of two procedures are as follows:

\begin{table*}[t]
	\centering
	\caption{Quantitative results on ISIC3 subset dataset show the superiority of KANResDiff. \textbf{Bold} represents the best per column.}
	\label{tab:2}
	\begin{tabular}{l|ccc|c|c}          			
		\toprule          			
		Methods & $\text{GED}_{16} \downarrow$ & $\text{GED}_{32} \downarrow$ & $\text{GED}_{100} \downarrow$ &  $\text{HM-IoU}_{32} \uparrow$ & $\text{MDM}_{32}$ $\uparrow$ \\
		\midrule      		
		Prob. Unet\cite{kohl2018probabilistic}  & $0.202_{\pm0.003}$ & $0.187_{\pm0.003}$ &  $0.171_{\pm0.002}$ & $0.697_{\pm0.005}$ &  $0.927_{\pm0.003}$\\ 
		SSN\cite{monteiro2020stochastic} & $0.197_{\pm0.007}$ & $0.181_{\pm0.004}$  &  $0.167_{\pm0.002}$ & $0.700_{\pm0.004}$ & $0.939_{\pm0.001}$\\ 			
		c-Prob. Unet\cite{17} & $0.208_{\pm0.004}$ & $0.202_{\pm0.005}$  & $0.179_{\pm0.002}$ &  $0.719_{\pm0.004}$ & $0.925_{\pm0.004}$\\ 			
		c-SSN\cite{17} & $0.200_{\pm0.007}$ & $0.195_{\pm0.007}$  & $0.177_{\pm0.001}$  & $0.725_{\pm0.004}$ & $0.931_{\pm0.003}$\\ 		
		MoSE\cite{20} &  $0.304_{\pm0.003}$ &  $0.275_{\pm0.004}$  & $0.232_{\pm0.001}$  & $0.673_{\pm0.001}$ & -\\ 
		ContourMS\cite{wang2025contour} & $0.223_{\pm0.003}$ & $0.202_{\pm0.002}$ & $0.174_{\pm0.001}$ & $0.764_{\pm0.005}$ & - \\
		KANResDiff(Ours) & $\mathbf{0.160_{\pm0.003}}$ & $\mathbf{0.151_{\pm0.002}}$ & $\mathbf{0.139_{\pm0.003}}$ & $\mathbf{0.773_{\pm0.002}}$ & $\mathbf{0.942_{\pm0.003}}$ \\  
		\bottomrule
	\end{tabular} 
\end{table*}

\noindent\textbf{Forward Procedure:} For the forward procedure, we aim to achieve a Schr\"odinger Bridge diffusion path. Therefore, $h_{\mathrm{RSB}}(t)$ is optimized to matching the control drift to the SB correction term:
\begin{equation}
	\mathcal{L}_{\mathrm{RSB}}^{(2n+1)}
	=
	\mathbb{E}_{t,x_t}
	\left[
	\left\|
	u_\phi(x_t,t)
	-
	\beta(t)\nabla_x \log \Psi(x_t,t)
	\right\|^2
	\right].
\end{equation}

\noindent\textbf{Backward Procedure:} For the backward procedure, we aim to update the predictions with the latest diffusion path. Hence, the magnitude of the control drift is regularized to ensure minimal deviation from the pretrained diffusion:
\begin{equation}
	\mathcal{L}_{\mathrm{RSB}}^{(2n+2)}
	=
	\mathbb{E}_{t,x_t}
	\left[
	\left\|
	\nabla_x \log p_t^{\mathrm{SB}}(x_t)
	-
	\nabla_x \log p_t^{\mathrm{data}}(x_t)
	\right\|^2
	\right].
\end{equation}

\section{Experiments}

\subsection{Experimental Setup.}
\noindent\textbf{Datasets.} Two public datasets are selected for evaluation: LIDC-IDRI \cite{23} and ISIC3 subset \cite{24,17}. LIDC dataset consists of lung CT scans with four corresponding segmentation labels, preprocessed following \cite{kohl2018probabilistic,16}. ISIC3 subset from \cite{17} provides dermoscopic images of skin lesions with three corresponding labels.

\noindent\textbf{Implementation Details.}
All training and inference procedures are conducted on a single RTX 4090 GPU. We set T=1000 and a batchsize of 8 for all experiments. Models are optimized using an Adam optimizer \cite{28} with a learning rate of $10^{-4}$. ITE and RSB are both trained for 600 epochs. 

\noindent\textbf{Evaluation Metrics.}
Generalized Energy Distance (GED) \cite{18}, Hungarian Matching IoU (HM-IoU) \cite{20}, and Maximum Dice Matching (MDM) \cite{rahman2023ambiguous} are adopted for evaluation. These metrics assess distributional alignment, overall matching accuracy, and maximum individual performance. Subscript n denotes the use of n samples. All results are reported as mean ± standard deviation over five runs.
\subsection{Experimental Results.}
\noindent\textbf{Quantitative Evaluations.} Table \ref{tab:1} and \ref{tab:2} report the quantitative comparisons among KANResDiff and advanced methods on LIDC dataset and ISIC subset. KANResDiff consistently achieves the best performance by reducing $\text{GED}_{16}$, $\text{GED}_{32}$, and $\text{GED}_{100}$ by a maximum of $18.8\%$, $16.6\%$, and $16.8\%$, respectively, indicating that the proposed dynamic and progressive stochastic modeling enables more effective ambiguity representation under all sampling strategies. A maximum of $7.7\%$ improvement in HM-IoU$_{32}$ further demonstrates the superior fidelity of the generated samples from KANResDiff. Meanwhile, KANResDiff maintains MDM performance comparable to supervised sample-level segmentation methods, demonstrating competitive individual accuracy.

\noindent\textbf{Qualitative Evaluations.} Fig. \ref{fig:pred} illustrates the segmentation visualization comparison across ground truths, the proposed KANResDiff and advanced methods. Predictions from KANResDiff better meet the pattern with ground truths compared with other methods, further validating its superiority.

\noindent\textbf{Ablation Study.} Table \ref{tab:3} shows the ablation study on LIDC dataset evaluating four model variants: ResDiff baseline, ResDiff with ITE, ResDiff with RSB and KANResDiff. Introducing ITE and RSB both lead to consistent improvements across all metrics, indicating the effectiveness of innovative components. Performance is further boosted with both components are jointly applied, which further verifies that ITE and RSB are complementary and that their integration effectively improves both segmentation accuracy and diversity.

\begin{table*}[t]
	\centering
	\caption{Ablation study on LIDC dataset demonstrates effectiveness of innovative components. \textbf{Bold} represents the best per column.}
	\label{tab:3}
	\begin{tabular}{l|cc|c|c|c}          			
		\toprule          			
		Methods & ITE & RSB & $\text{GED}_{100} \downarrow$ &  $\text{HM-IoU}_{32} \uparrow$ & $\text{MDM}_{32}$ $\uparrow$ \\
		\midrule      		
		Baseline (ResDiff) & $\times$ & $\times$ & $0.187_{\pm0.002}$ & $0.667_{\pm0.002}$ & $0.903_{\pm0.003}$ \\
		ResDiff with ITE & $\checkmark$ & $\times$ & $0.181_{\pm0.003}$ & $0.683_{\pm0.001}$ & $0.908_{\pm0.002}$ \\
		ResDiff with RSB & $\times$ & $\checkmark$ & $0.174_{\pm0.004}$ & $0.688_{\pm0.004}$ & $0.916_{\pm0.002}$ \\
		KANResDiff & $\checkmark$ & $\checkmark$ & $\mathbf{0.172_{\pm0.001}}$ & $\mathbf{0.701_{\pm0.001}}$ & $\mathbf{0.930_{\pm0.002}}$ \\
		\bottomrule
	\end{tabular} 
\end{table*}

\section{Conclusion}
In this work, we proposed KANResDiff for learning local residual diffusion, promoting a flexible stochastic intensity instead of predefined injection for ambiguous medical image segmentation. Specifically, KANResDiff offers a progressive, dedicated ambiguity-aware semantic modeling process by assigning independent and local residual-based diffusion formulation across stages. The proposed Independent Time Encoding offers spline-based time embeddings that can be locally optimized, enhancing independence across stages. The proposed Residual Schr\"odinger Bridge introduces deterministic residual via learnable weights by constructing local optimal diffusion path with Schr\"odinger Bridge, forming a flexible determinism coordination with stochasticity in diffusion models. Experimental results on two public datasets demonstrate our KANResDiff outperforms existing methods. Through multifaceted analysis and comprehensive evaluation, our KANResDiff offers a promising approach for ambiguous medical image segmentation that eliminates excessive or inadequate stochasticity.

\begin{credits}
\subsubsection{\ackname} This work was supported by the National Natural Science Foundation of China under Grant No. 62501195; the Shenzhen Medical Research Fund under Grant No. C2501016; the National Natural Science Foundation of China under Grant No. 82527807; the Science and Technology Innovation Committee of Shenzhen Municipality under Grant No. JCYJ20250604145426036; the Key Research \& Development Program of Heilongjiang Province under Grant Nos. 2024ZX12C23 and 2023ZX01A08; and the Natural Science Foundation of Heilongjiang Province under Grant No. LH2024F019.

\subsubsection{\discintname}
The authors have no competing interests to declare that are relevant to the content of this article.
\end{credits}
%
%
%
%
\bibliographystyle{splncs04}
\bibliography{Paper-0621}
\end{document}